\documentclass[letterpaper, 10 pt, conference]{ieeeconf}  

\IEEEoverridecommandlockouts                              

\title{\LARGE \bf
Graph Neural Network Based Method for Path Planning Problem 
}

\author{Xingrong Diao, Wenzheng Chi*, \textit{Senior Member, IEEE}, Jiankun Wang*, \textit{Senior Member, IEEE}
\thanks{This work is supported by Shenzhen Outstanding Scientific and Technological Innovation Talents Training Project under Grant RCBS20221008093305007, and National Natural Science Foundation of China grant \#62103181 (\textit{Corresponding authors: Wenzheng Chi, Jiankun Wang}).}
\thanks{Xingrong Diao and Jiankun Wang are with Shenzhen Key Laboratory of Robotics Perception and Intelligence, Department of Electronic and Electrical Engineering, Southern University of Science and Technology, Shenzhen 518055, China (e-mail: {\tt\small 12332163@mail.sustech.edu.cn}; 
{\tt\small wangjk@sustech.edu.cn}).}%
\thanks{Jiankun Wang is also with Jiaxing Research Institute, Southern University of Science and Technology, Jiaxing, China.}
\thanks{Wenzheng Chi is with the Robotics and Microsystems Center, School of Mechanical and Electric Engineering, Soochow University, Suzhou 215021, China (e-mail: {\tt\small wzchi@suda.edu.cn})}
}

\usepackage{amsfonts,amssymb}
\usepackage{amsmath}
\usepackage{graphicx}
\usepackage{epstopdf}
\usepackage{subcaption}
\usepackage{cleveref}
\usepackage{lipsum}
\usepackage{wrapfig}
\usepackage[ruled,linesnumbered]{algorithm2e}

\begin{document}

\maketitle
\thispagestyle{empty}
\pagestyle{empty}

\begin{abstract}

Sampling-based path planning is a widely used method in robotics, particularly in high-dimensional state space. Among the whole process of path planning, collision detection is the most time-consuming operation. In this paper, we propose a learning-based path planning method that aims to reduce the number of collision detections. We develop an efficient neural network model based on Graph Neural Networks (GNN). The model outputs weights for each neighbor based on the obstacle, searched path, and random geometric graph, which are used to guide the planner in avoiding obstacles. We evaluate the proposed method's efficiency through simulated random worlds and real-world experiments, respectively. The results demonstrate that the proposed method significantly reduces the number of collision detections and improves the path planning speed in high-dimensional environments.
\end{abstract}
    
\section{INTRODUCTION}

The path planning problem in robotics is to find a collision-free path from the initial state and the goal state of a robot, given a description of the environment. In recent decades, graph-search and sampling-based methods have become two popular techniques for path planning problems in robotics. Graph-search methods, such as Dijkstra \cite{c1} and A* \cite{c2}, usually search in discrete space, and the quality of their solution is often related to the degree of discretization. However, as the dimension of configuration space grows, these methods often fall into the curse of dimensionality \cite{c3}, making them computationally difficult.
In contrast, sampling-based methods such as Probabilistic Roadmap (PRM) \cite{c4}, Rapidly-exploring Random Tree (RRT) \cite{c5}, and Expansive Space Trees (EST) \cite{c6} improve efficiency and scalability in high-dimensional spaces by avoiding discretization and explicit representation of the configuration space. They explore the whole space by random sampling, resulting in probabilistic completeness for the feasible solution. Some sampling-based methods use graph-search methods concepts to find the path, such as Fast Marching Trees (FMT*) \cite{c7} and Batch Informed Trees (BIT*) \cite{c8}. FMT* and BIT* use the heuristic function to sort the samples and edges to explore, which improves the initial solution and convergence rate to the optimum.

Many sampling-based methods are improved by modifying the sampling distribution, such as Gaussian PRM \cite{c9}, and GAN-Based heuristic RRT\* \cite{c10}. However, for most existing methods, collision detection is a major computational bottleneck because they need to repeatedly check the path to ensure that the path is collision-free. Typically, a path planning method spends about 70\% of the computation time on collision detection. Lazy PRM \cite{c22} reduces collision detection by checking the edge only when it is on the global shortest path. Although it is useful in high dimensions, it does not guarantee robustness.

To address the aforementioned issues, we propose a learning-based path planning method for reducing the number of collision detections. Our method uses a graph neural network (GNN) model to predict the edge weights of the neighbor set of the current vertex. The weights are used to guide the planner to avoid obstacles and accelerate the search process. We evaluate the proposed method in simulations and real-world experiments and obtain good performance. Compared with classical path planning problems, our method significantly improves the path planning speed, reduces the number of collision detection, and improves the success rate and robustness in a high-dimensional environment.

\begin{figure}[t]
    \centering
    \subfloat[]{\includegraphics[width=1.3in,height=1.3in]{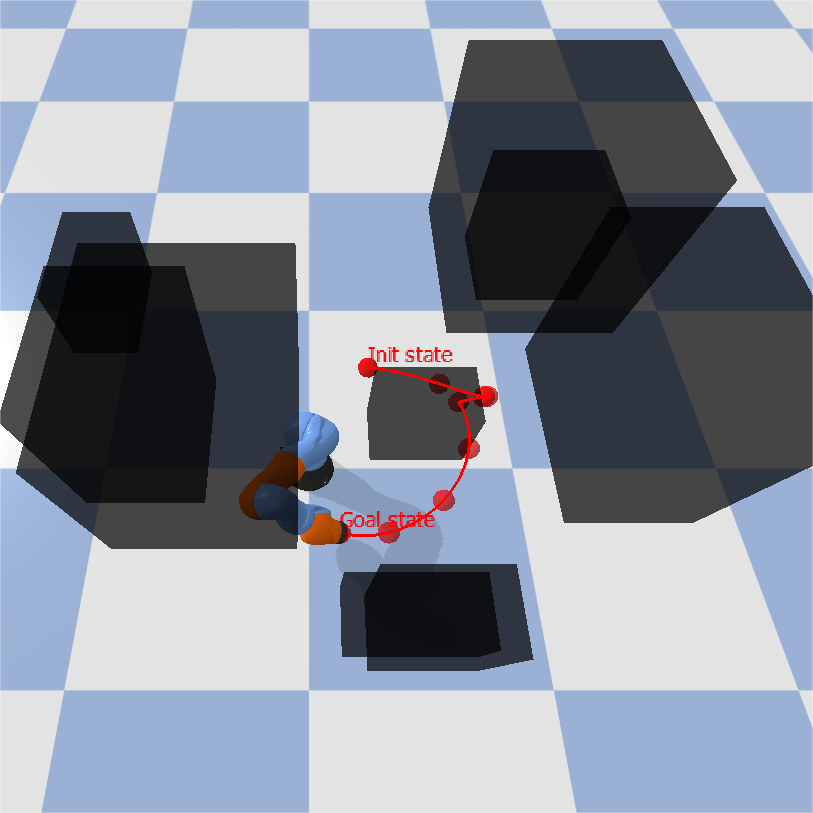}}
    \hspace{0.3in}
    \subfloat[]{\includegraphics[width=1.3in,height=1.3in]{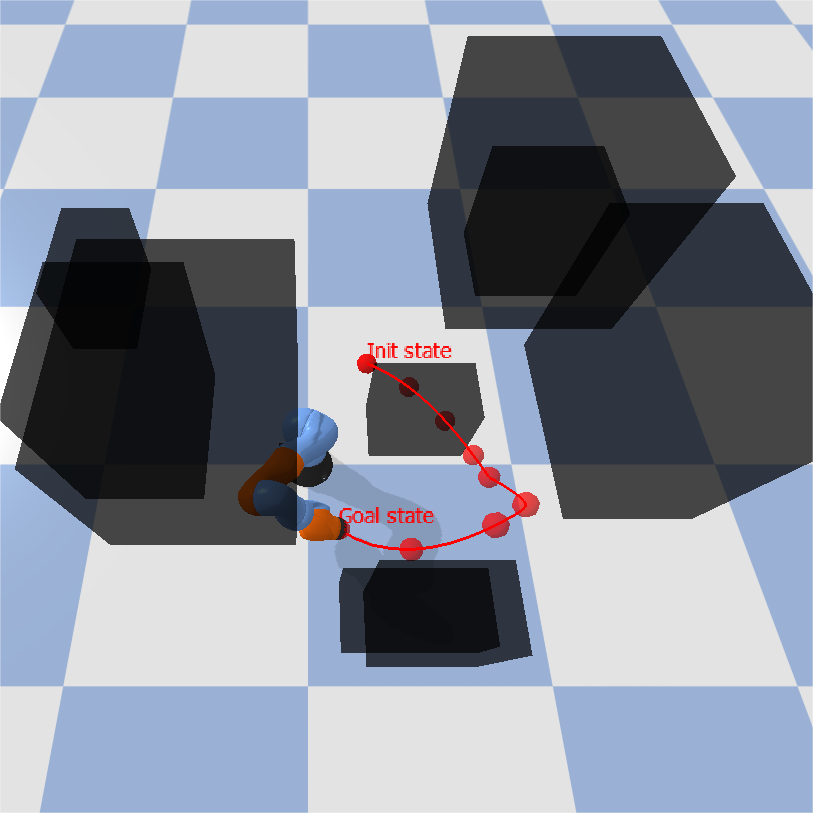}} 

    \caption{Demonstrations for our method in 7D environment. The collision check of our method is 0.94 times and 0.0049 times of BIT* and PRM, respectively. The planning time of our method is 1.4\% and 1.1\% of the BIT* and PRM, respectively. From Left to Right: (a) GNN model. (b) BIT*}
    \label{Fig1}
\end{figure}
Our main contributions include:

1) Propose a heuristic method for sampling-based path planning with GNN.

2) Design a GNN model to predict weights for each node in the neighbor set of the current vertex.

This paper is structured as follows: Section \ref{tit2} provides a comprehensive review of the previous work. In Section \ref{tit3}, we introduce some preliminaries about this work. The dataset construction and the proposed GNN model are described in Section \ref{tit4}. Subsequently, Section \ref{tit5} presents the experimental results. Finally, Section \ref{tit6} concludes the paper.

\section{RELATED WORK}
\label{tit2}
Recently, many studies have proposed different types of heuristics to improve the performance of motion planning. 

One direction is to obtain more valuable sampling points through bias sampling. Ichter et al.\cite{c20} propose a general methodology for sampling based on conditional variational autoencoder (CVAE) \cite{c18}. Zhang et al. \cite{c10} design a Generative Adversarial Network (GAN) \cite{c27} to generate a promising region that may contain feasible paths. Liu et al. \cite{c32} establish a set of partition heuristic rules for target and collision-free guidance to bias the search towards the target. Baldwin et al. \cite{c33} use expert data to learn sampling distributions, and an estimate of sample densities around semantic regions of interest, then incorporate them into a sampling-based planner to produce natural plans. The method of bias sampling improves the efficiency of sampling and the performance of path planning, but they are prone to fall into the local optimal solution and still need to spend a lot of calculations on collision detection.

Besides bias sampling, one of the other direction is lazy motion planning which focus on reducing collision detection. Lazy PRM \cite{c22} and Lazy SP \cite{c23} perform motion planning by generating a random geometric graph (RGG) and checking the edge only when it is on the global shortest path of the target. BIT* \cite{c8} uses a heuristic to efficiently search a series of increasingly dense implicit RGGs while reusing previous information, and it effectively reduces the number of collision checks through batch sampling and incremental search. Fast \cite{c34} and Clearance Net \cite{c35} perform collision detection by learning function approximators. However, most Lazy methods are hand-crafted, which is unsuitable for general planning problems.

For decades, as neural networks have evolved, many researchers have begun to use learning-based path planning. The Neural Exploration-Exploitation Trees (NEXT) algorithm \cite{c21} has demonstrated the clear benefits of using learning-based components to reduce samples and accelerate programming. L2RRT \cite{c25} first embedded high-dimensional structures into low-dimensional representations, and then used RRT for motion planning. However, most of the existing learning-based planning algorithms use Convolutional Neural Networks (CNN), like the neural RRT*\cite{c26}, to learn the planning environment, which loses the structure of the environment through the sampling points. 

Graph Neural Networks (GNNs) have solved the problem with graph structure such as text classification \cite{c29}, protein interface prediction \cite{c30}, and parsing social relationships \cite{c31}. Yu et al. \cite{c24} use GNN to reduce the number of collision checking times on the edges of PRM by ignoring the edges that may interfere with the obstacles. Li et al. \cite{c28} use CNN to extract adequate features from local observations, with GNN to communicate the features among robots to realize Multi-Robot path planning. Khan et al. \cite{c19} analyze the feasibility of using GNN to perform classical motion planning problems, and use GNN to identify critical nodes or learn the sampling distribution in RRT. Our work focuses on lazy planning and learning-based planning and aims to design a GNN-based model for reducing collision detection to improve the performance of the sampling-based method.

\section{PRELIMINARIES}
\label{tit3}
\subsection{Feasible Planning}
 We define the optimal planning problem similarly to \cite{c11}. Let $X\subseteq \mathbb{R}^n$ be the state space of the planning problem, where $n \subseteq \mathbb{N}$ is the dimension of the state space, $X_{obs}\subset X$ be the obstacle space, and $X_{free} = X/X_{obs}$ be the free space. Let $x_{init}\in X_{free}$ be the initial state and $X_{goal}\subset X_{free}, X_{goal}=\{x\ |\ D(x_{goal},x)<\Delta \}$ be the goal region, which $D(\cdot)$ is Euclidean distance and $\Delta$ means tiny numbers. Let $\sigma:[0,1]\rightarrow X_{free}$ be a feasible path of the planning problem and $c(\sigma)$ be the cost function. The solution is the optimal path $\sigma ^{*}$, which minimizes the cost function $c:\Sigma \rightarrow \mathbb{R}_{\geq 0}$, where $\mathbb{R}_{\geq 0}$ is the set of non-negative numbers,
\begin{equation}
\begin{aligned}
    \sigma ^{*} =& \mathop{\rm arg\ min} \limits_{\sigma\in \Sigma} c(\sigma)\\ 
    s.t.\ &\sigma(0) = x_{init},\\
    &\sigma(1)\in X_{goal},\\
    &\sigma(t)\in X_{free},\ \forall t\in [0,1].
\end{aligned}
\end{equation}
\subsection{Random Geometric Graph}
A random geometric graph (RGG) consists of a node set $\mathcal{N}$ and an edge set $E$. The node set consists of nodes (states) sampled from the state space, while the edges set is constructed with the edge connected by two states based on the relative geometric position of the two states. When an edge connects two states, the two states are neighbors of each other. A neighbour set $\mathcal{N}_x$ of a state $x$ is defined as follow:
\begin{equation}
\begin{aligned}
    \mathcal{N}_x = \{\tilde{x}\ |\ &D(x,\tilde{x})\leq d_{max},\\
    &D(x,\tilde{x})=||x-\tilde{x}||\},
\end{aligned}
\end{equation}
where $d_{max}$ is the constant threshold of the relative distance, $||x-\tilde{x}||$ is the Euclidean distance between $x$ and $\tilde{x}$. Common RGGs have a specific number of edges of neighbors closest to each state (a k-nearest graph \cite{c12}) or to all neighbors at a specific distance (an r-disc graph \cite{c13}). We use the r-disc graph in the proposed method, which is defined as $G=\{\mathcal{N},E\}$, where $\mathcal{N}=\{x\ |\ x\sim U(X_{free})\}$ is the states set, with the uniform distribution $U(\cdot)$, and $E=\{(x,w)\ |\ D(x,w)<r\}$, $x,w\in \mathcal{N}$, $r\in \mathbb{R}_{\geq0}$ is the edges set. In the proposed method, the $x_{init}$ and $x_{goal}$ are added to the state set before the edges are connected, which means $x_{init},x_{goal}\in \mathcal{N}$.

\subsection{GNN-based planner}
Sampling-based planners can be viewed as methods that construct an implicit RGG and explicit spanning tree in the free space of the planning problem, which means that the performance of the planners depends on the quality of the RGG and the efficiency of the spanning tree. However, most sampling-based planners spend about 70\% of the time on collision detection to keep the spanning tree in free space. Therefore, to improve the efficiency of the spanning tree, we construct a GNN model to reduce the number of collision detections.

\begin{algorithm} \footnotesize
\label{algo}
    \caption{GNN-based planner} 
        \KwData{Problem $U=\{G=\{\mathcal{N},E\}, O,x_{init},x_{goal}\}$\;}
        \KwResult{path $\sigma$\;}
        $Flag(x_i)=0,\ \forall x_i\in \mathcal{N}$, $Flag(x_{init})=1$\;
        $\sigma\leftarrow x_{init}$, $x_{cur}=\sigma(0)$\;
        \While {Ture}{
            $\Omega =model(\sigma,G,O)$\;
            $rank = Sort(\Omega,x_i)$\;
            \For{$x_i \in rank$}{
                \If{$CollisionFree(x_i,\ x_{cur})$}{
                    $\sigma\leftarrow \sigma\cup x_i$\;
                    $x_{cur}= x_i$\;
                    break\;
                    }                
                $rank \Leftarrow rank/x_i$\;
                }
            \If{$rank = \emptyset$}
            {return False\;}
            \If{$x_{goal}\in \sigma$}
            {break\;}
        }
        \Return $\sigma$\;
\end{algorithm}

\begin{figure}[h]
\centerline{
\includegraphics[totalheight=1.5in]{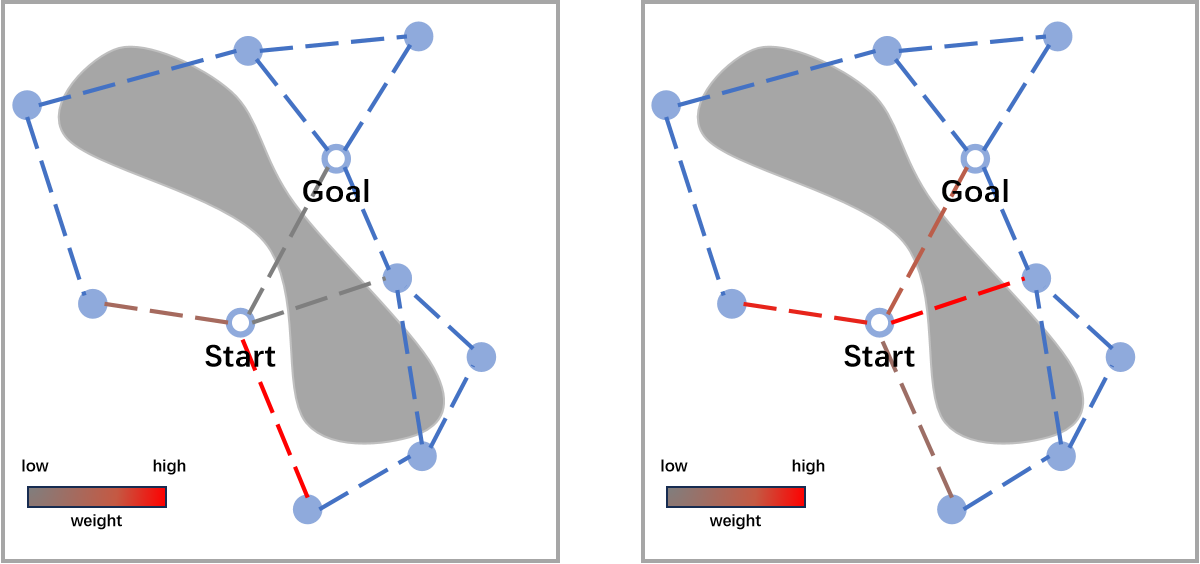}}
\caption{Difference between GNN and PRM when selecting a new state. The redder the color of an edge, the higher the guidance value of the neighbor. Left: GNN model. Right: PRM.}
\label{Fig2}
\end{figure}

To solve a path planning problem, the proposed method first samples a batch of states, and adds $x_{init}$ and $x_{goal}$ to the batch of states, with edge connections to build an RGG. Then, the GNN model takes $x_{init}$ as the first vertex of the path $\sigma(0)$ and outputs a value set, which is called the guidance value set and explained in Section \ref{tit4}, for the neighbors of $\sigma(0)$. The method first selects a neighbor with a greedy algorithm and then checks whether the edge between the neighbor and the current state is collision-free through the simulation environment. The method will ignore the neighbor if it is not collision-free and uses the greedy algorithm to select the next neighbor in descending order according to the guidance value. In such order, a collision-free neighbor will added as a new vertex of the searched path. The search will fail if no neighbor is collision-free. The equation of the greedy algorithm is defined as: 
\begin{equation}
\begin{aligned}
\sigma_{k+1}= \mathop{\rm argmax}\limits_{\omega_i}\ {\rm Model}(\sigma_k|\mathcal{N},E,O),
\end{aligned}
\end{equation}
where $\sigma_k$ is the last vertex of the path that has been searched, ${\rm Model}(\cdot)$ means the output of the GNN-based model, $\omega_i$ means the guidance value of the neighbor of $\sigma_k$, and $O$ is the obstacles set. We call such a step model-greedy-detection, which means that firstly, the model outputs the evaluation of each neighbor, then selects a neighbor through the greedy algorithm, and finally does collision detection. The model-greedy-detection is shown in Alg. 1, Line 5-14. The proposed method will repeat model-greedy-detection on the last vertex $\sigma(k)$ of the searched path $\sigma(t),\ \forall t\in[0,k],\ k<1$ for getting new vertices. The method ends the loop until a path from $x_{init}$ to $x_{goal}$ is found and outputs a path as the solution to the problem. In other words, the method finds the path without developing a spanning tree. Fig. \ref{Fig2} shows the difference between our method and PRM when selecting a new state.

\subsection{Probabilistic Completeness} 
Probabilistic completeness means that as the number of samples goes to infinite, the probability of finding a feasible solution equals 1 if it exists. Because the proposed method constructs an RGG by sampling uniformly in free space, when the sampling states in RGG go to infinity, the method will be probabilistic complete, as shown in the following equation:
\begin{equation}
\begin{aligned}
    \mathop{\rm lim}\limits_{{\rm Num}\rightarrow \infty} \mathbb{P}(\mathcal{N}\cap X_{goal}\neq \emptyset)=1.
\end{aligned}
\end{equation}

In other words, if there is a feasible solution, it will be found as the number of sampling states tends to infinity, which guarantees the probabilistic completeness of the method.

\section{GNN-BASED METHOD FOR PATH PLANNING}
\label{tit4}
In this section, we introduce the proposed GNN model in detail. The GNN model is trained with a large amount of planning cases with conditions consisting of initial state $x_{init}$, goal state $x_{goal}$, an RGG $G$, and obstacles $O$. The model will synthesize the information of obstacles, searched path, and goal state, and output a guidance value set. The guidance value is the model's evaluation of each neighbor for the last vertex of the searched path, which is used to guide the greedy algorithm to search for a feasible path. The larger the guidance value is, the more likely the model considers the corresponding neighbor to be collision-free and on the optimal path. We sign the guidance value as $\omega_i$, and the guidance value set as $\Omega$. Each state $x\in \mathcal{N}$ has its own guidance value set, which changes with the search of the path. 

\subsection{Dataset Generation}
\label{subtit41}
\begin{figure}[h]
\centerline{
\includegraphics[width=3in]{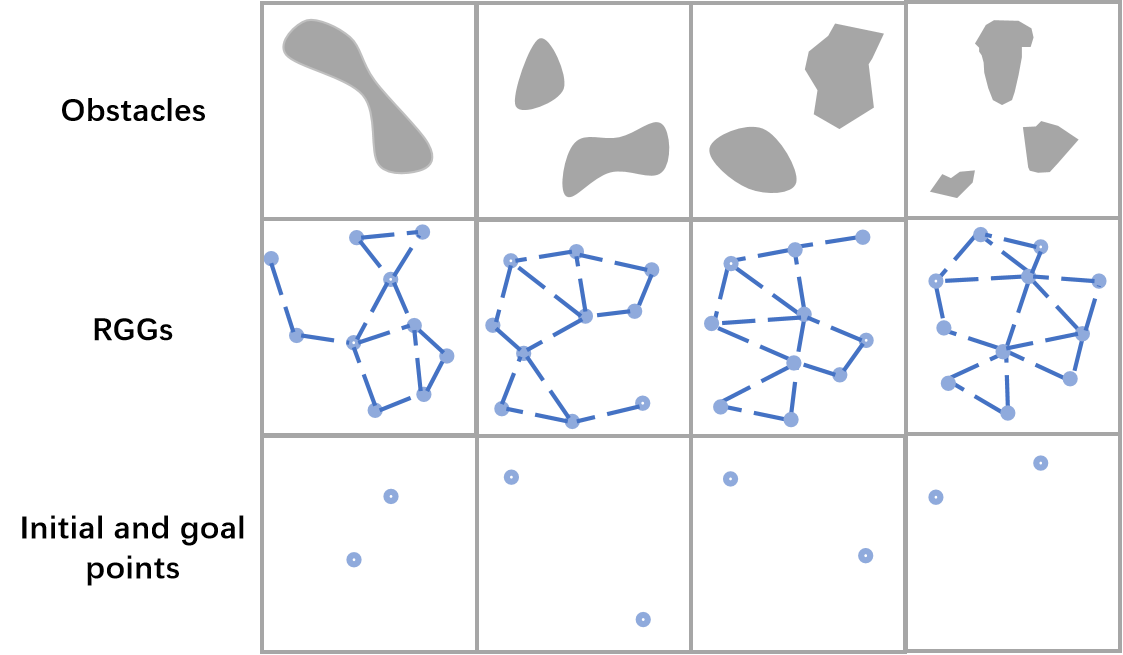}}
\caption{Schematic diagram of the dataset. Each column from top to bottom represents obstacles, RGGs, and initial and goal states.}
\label{Fig8}
\end{figure}

As shown in Fig. \ref{Fig8}, each dataset consists of an initial state $x_{init}$, a goal state $x_{goal}$, a set of obstacles $O$, and an RGG. It should be noted that Fig. 3 is a schematic diagram of the training set, in the real training set, all random points in RGG are $n$ dimensional arrays, and obstacles are represented by 6-dimensional arrays. The dataset does not contain feasible paths, which are generated as the training process progresses. We use a 6-dimensional array to represent an obstacle, which represents centroid coordinates and the length, width, and height of the obstacle, respectively. Based on that, placing $m$ obstacles in a simulator requires $m$ 6-dimensional arrays, where $m$ is a random integer, and the 6-dimensional arrays are uniform sampling from the workspace of the robotic arm, with limitation of the size (low: 0.1, high: 0.3), and do not interface with the robot arm when the state of the arm is $ \overrightarrow{0}\in \mathbb{R}^n$. The RGG $G=\{\mathcal{N}, E\}$ is built by randomly sampling in the workspace of the robotic arm, with edge connections based on the r-disc rule. The data of each state in RGG is shown in the next part, and each state has no label, the same as $x_{init}$ and $x_{goal}$.

\begin{figure*}[t]
\centerline{
\includegraphics[totalheight=2.3in]{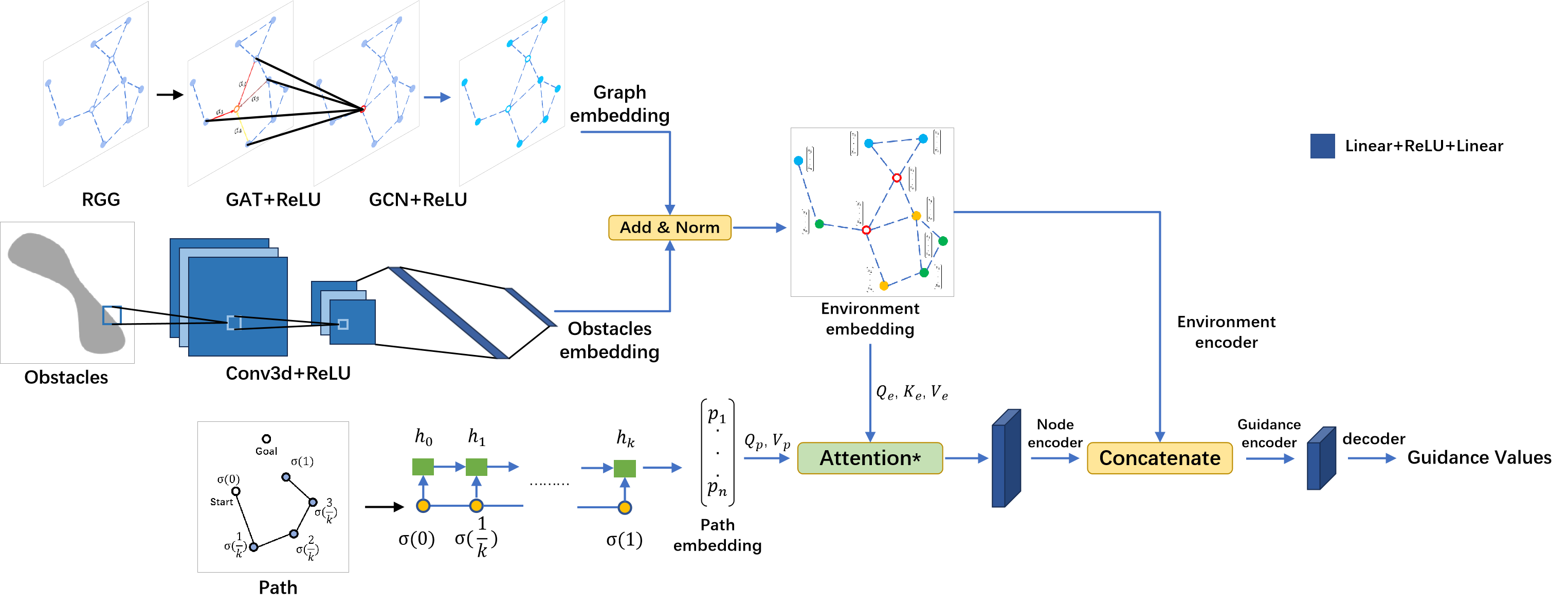}}
\caption{The detailed structure of the GNN model for calculating guidance value set. The ${\rm Attention*}=$${\rm Norm}({\rm Attention}([Q_e K_e^T\ |\ Q_p K_e^T],\ [V_e\ |\ V_p]),\ {\rm \mathcal{O}})$, where $[\cdot|\cdot]$ means concatenation, and Norm is Layer Normalization \cite{c36}.}
\label{Fig3}
\end{figure*}

\subsection{Model Structure}
\subsubsection{Overview}
We mainly use GNNs, recurrent neural network (RNN) \cite{c16}, and attention mechanism \cite{c17} to construct our model. GNNs are mainly used to process data with graph structure, such as protein structure, social network, and network connection relationships. It aggregates the information in the graph without changing the structure of the graph. By aggregating information about surrounding neighbors, the nodes in the graph get a richer representation. As sampling-based methods use implicit graphs to learn the information of the free space, using GNNs to aggregate the information can enable planners to obtain global information and improve planning capabilities. Based on the definition in Section \ref{tit3}, a finite graph where each state $x_i$ has data $h_i^0\in\mathbb{R}^{2n}$, and a typical GNN encodes the representation $h^{k+1}_i$ of the state $ x_i$ after k aggregations as:
\begin{equation}
\centering
\begin{aligned}
    c_i^{k}=\oplus(\{f(x_i^k,&x_j^k\ |\ (x_i,x_j)\in E)\}),\\
    h_i^{k+1}&=g(h_i^k,c_i^k),
\end{aligned}
\end{equation}
where $f$ and $g$ are fully connected networks and $\oplus$ is an aggregation function on sets, such as mean and sum. In the proposed model, we use Graph Convolutional Network (GCN) \cite{c14} and Graph Attention Network (GAT) \cite{c15} to obtain node features in finite graphs, where the function for GCN is defined as:
\begin{equation}
\begin{aligned}
    H^{k+1}=\sigma(\tilde{D}^{-\frac{1}{2}}\tilde{A}\tilde{D}^{-\frac{1}{2}}H^kW^k),
\end{aligned}
\end{equation}
where $H^k$ is the data set of all nodes in $k^{\rm th}$ aggregation, $\tilde{A}=A+I$ ( $A$ is the adjacency matrix, $I$ is the identity matrix), $\tilde{D}$ is the degree matrix of $\tilde{A}$ ($\tilde{D}_{ii}=\sum j\tilde{A}$), $\sigma$ is a nonlinear activation function (in our method, $\sigma$ is ReLU). Also, the function of GAT is defined as:
\begin{equation}
\begin{aligned}
    h_i^{'}&=\sigma (\sum_{j\in N_i}\alpha_{ij}Wh_j),\\
    \alpha_{ij}&=\frac{\exp({\rm LeakyReLU}(e_{ij}))}{\sum_{k\in N_i}a_{ij}Wh_j},\\
    e_{ij}&=a([Wh_i\ ||\ Wh_j]),\ j\in \mathcal{N}_i,
\end{aligned}
\end{equation}
where $\mathcal{N}_i$ is the neighbors set of a state $x_i, x_i\in \mathcal{N}$, $a$, and $W$ are the weights of the model, $\alpha_{ij}$ is the weight of each edge connected to $x_i$, and ${\rm LeakyReLU}$ is an activation function. The data $h_i$ of the state $x_i$ is constructed by concatenating $x_i$ and $x_{goal}$, which is written as $h_i=[x_i,\ x_{goal}]$. Besides, given GCN $C$ and CAT $T$, the node embedding of GNN in our model is calculated as:
\begin{equation}
\begin{aligned}
    x=\sigma C(\sigma T(h_i^k)).
\end{aligned}
\label{attenofmodel}
\end{equation}

In our model, GAT is used to preliminarily determine the priority of neighbors (Euclidean distance and dispersion), and GCN is used to collect the connections of second-order neighbors. We also use an attention mechanism to encode the environment and the path together. Given $n$ keys with dimension $d_k:K\in\mathbb{R}^{n\times d_k}$, the value that corresponds to the key $V\in \mathbb{R}^{n\times d_v}$, and $m$ query values $Q\in \mathbb{R}^{m\times d_q}$, we can encode them with attention mechanism as:
\begin{equation}
\begin{aligned}
    {\rm Attention}(Q,K,V)={\rm Softmax}(\frac{QK^T}{\sqrt{d_k}})V.
\end{aligned}
\end{equation}

\subsubsection{Detailed Model Structure}
\label{subtit422}
\begin{figure}[h]
\centerline{
\includegraphics[totalheight=1.5in]{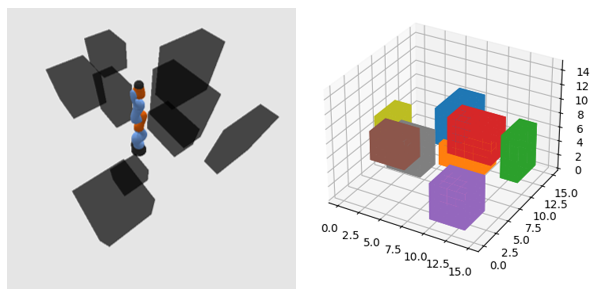}}
\caption{Comparison of the obstacles in simulation world and mapping space. Left: Simulation world. Right: Mapping space.}
\label{Fig4}
\end{figure}

The proposed model mainly comprises three parts: environment encoder, path encoder, and decoder. The environment encodes the obstacles and the sampling points. Based on that, we record the path information through the path encoder to avoid duplicate searches. The outputs of the two encoders will be aggregated by the attention mechanism and decoded by the decoder to obtain the guidance value set.
Fig. \ref{Fig3} illustrates the detailed structure of the proposed model. Before running the model, the obstacle arrays need to be mapped in $\mathbb{R}^{m\times m\times m}$ by a mapping function $f_m:\mathbb{R}^6\to\mathbb{R}^{m\times m\times m}$, where $m$ is the size of the mapped space. The mapped space needs to reflect the conditions of the workspace, like the relative positions and sizes of the obstacles and is used to obtain the obstacles feature. The comparison of the obstacles in the simulation world and mapping space is shown in Fig. \ref{Fig4}.

In the environment encoder, the states are embedded by (\ref{attenofmodel}), with $x\in \mathbb{R}^{|X|\times d}$, and the obstacles are embedded by 3D CNN and linear network, with $ y\in \mathbb{R}^{d}$, where $d$ is the embedding size of states and obstacles. After that, the environment embedding is computed as $\mathcal{O}=g(x_i, y\ |\ e_i:(x_i,x_{j})),\ \forall x_i, x_j\in \mathcal{N},\ \mathcal{E}\in \mathbb{R}^{|X|\times d_o}$, where $g$ is two-layer MLP with normalization, $d_o$ is the embedding size of the environment. 

In the path encoder, the path is embedded by $\mathcal{P}=f_p(\sigma_{model})$, where $f_p$ is RNN, and $\sigma_{model}$ is the path that the method has searched. We then use the attention mechanism to encode the node embedding by $x_i={\rm Attention*}(Q_e,Q_p,K_e,V_p,V_e)$, where $Q_e, K_e$, and $V_e$ are calculated from $\mathcal{O}$ and $Q_p,V_p$ are calculated from $\mathcal{P}$. Next, the node embedding will be decoded as $x_i=h_x(x_i,x_i-x_j, o_i|\ e_i:(x_i,x_{j}),o_i\in \mathcal{O})$, where $h_x$ is two-layer MLP and $x_i,x_j\in N$. At last, the guidance value sets are calculated by two-layer MLP.

\subsubsection{Training Process}
Each training case consists of an initial state $x_{init}$, a goal state $x_{goal}$, a set of obstacles $O$, and an RGG based on r-disc rules. The goal is to train a GNN-based model to generate the guidance value set. Besides, if the model is trained with a reference path, there is no guarantee that the model will still work properly when it falls into some abnormal conditions, such as deviations from the optimal path or improper evaluation of some neighbors. To achieve better robustness, we use a formula, which is similar to $\sigma_{total} = \sigma_{model}(i)+\sigma_{Dijkstra}(j),$$\ i,\ j\in[0,1]$, where $\sigma_{model}(0)=x_{init},$$\ \sigma_{Dijkstra}(0)=\sigma_{model}(1), \sigma_{Dijkstra}(1)=x_{goal}$, to train the GNN model instead directly use the reference path computed from the Dijkstra algorithm.  Since our method only outputs paths without spanning trees, it is more appropriate to use the training method we mentioned.

In each training case, we first let the method output a path $\sigma_{model}$ of length $k\in Z$, which is randomly sampled from 1 to 10. The method then uses the GNN model to calculate the guidance value set for the neighbors of the last vertex $\sigma_{model}(1)$ of the path $\sigma_{model}$. After that, the Dijkstra algorithm is used to search a collision-free path $\sigma_{Dijkstra}$ from $\sigma_{model}(1)$ to $x_{goal}$, and we assume the length of $\sigma_{Dijkstra}$ is $j$. It then calculates the loss function, which is shown in the next part, by the guidance value set $\Omega$ of $\sigma_{model}(1)$ and the guidance value $\omega_i,\ \omega_i\in \Omega$, of the second vertex $\sigma_{Dijkstra}(\frac{1}{j})$ of the path $\sigma_{Dijkstra}$. Finally, we update the model's parameters through stochastic gradient descent (SGD), with learning rate = 0.001 and weight decay = 0.001.

\begin{figure*}[t]
\centerline{
\includegraphics[width=6.7in]{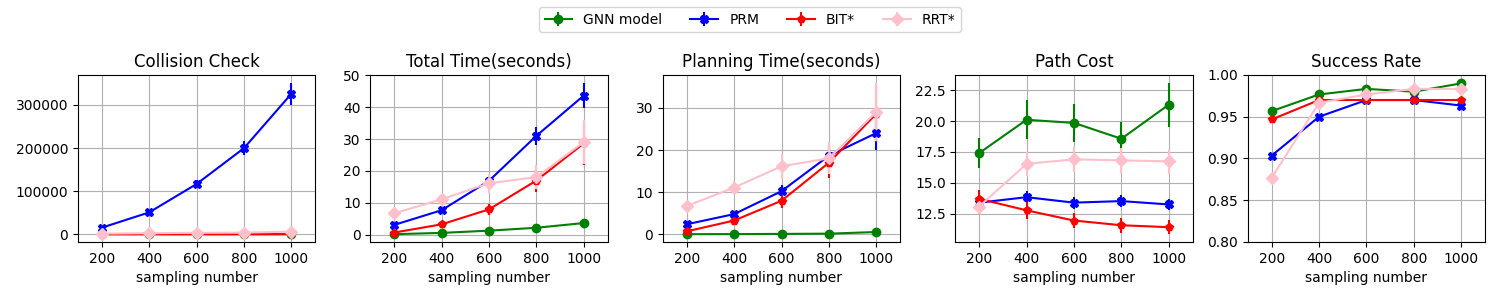}}
\caption{Comparison of performance in $\mathbb{R}^7$ at different sampling numbers. From left to right is: (a) Collision Check. (b) Total Time. (c) Planning Time. (d) Path Cost. (e) Success Rate.}
\label{Fig5}
\end{figure*}

\subsubsection{Loss Function}
The cross-entropy is used as the loss function of training, which can measure the difference between two distributions. The function of the cross-entropy can be defined as:
\begin{equation}
\begin{aligned}
    H(P,F)=&\mathbb{E}_{x\sim F}[-{\rm log} P(x)]\\
    =&-\sum\limits_{i=1}^nP(x_i){\rm log}F(x_i),
\end{aligned}
\end{equation}
where $P$ is the real distribution, and $F$ is the generated distribution. In our model, we refer to the distribution that gives the optimal path as $P$ and the distribution generated by the model as $F$. We calculate $x_i$ by the following equation:
\begin{equation}
    x_i={\rm Softmax(\omega _i)}=\frac{e^{\omega _i}}{\sum_{\omega_j\in\Omega}e^{\omega _j}},
\end{equation}
where $\omega _i$ represents the guidance value, $\Omega$ represents the guidance value set of the state $x_i$. Therefore, the loss function of the training process is defined as:
\begin{equation}
\begin{aligned}
    &\mathcal{L}(\omega,\ \Omega)=\mathbb{E}_{x\sim F}[-{\rm log} P(x)]\\
    &=-\sum\limits_{i=1}^nP(\frac{e^{\omega_i}}{\sum_{\omega_j \in \Omega}e^{\omega _j}}){\rm log}F(\frac{e^{\omega _i}}{\sum_{\omega_j \in\Omega}e^{\omega _j}}).
\end{aligned}
\end{equation}

\section{EXPERIMENTS}
\label{tit5}
Our dataset contains a total of 3300 sets of data. We choose 3000 of them for the training set and 300 for the test set, and Section \ref{subtit41} describes how the dataset was generated. In the experiment, we use Pybullet as a simulation environment to test the proposed method. Besides, we also test the feasibility of the path in real-world. We experimentally test the method with the number of collision detection, planning time, and cost of the path in the 7D-Kuka arm in simulated random worlds and the 6D-Kinova arm in real-world manipulation problems.

\begin{figure*}[t]
    \centerline{
    \includegraphics[width=7in]{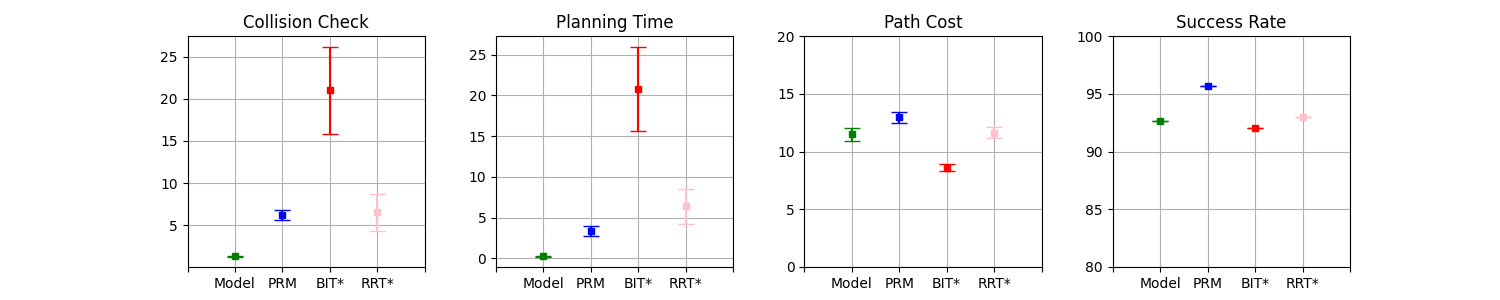}}
    \caption{Comparison of performance in $\mathbb{R}^6$ with sampling number = 400. From left to right is: (a) Collision Check. (b) Planning Time. (c) Path Cost. (d) Success Rate.}
    \label{Fig7}
\end{figure*}
\subsection{Simulated Random Worlds}
The proposed method is compared to existing sampling-based methods on random problems in $\mathbb{R}^7$. In addition to recording the number of collision detection, planning time, success rate, and cost of path. In each test, we use different numbers ($200,400,600,800,1000)$ of sampling points to observe the performance of each method. We also test the time in both the total time and the time without building RGG. Besides, the cost of the path is calculated by $c(\sigma)=\sum_{i=0}^{l-1} D(x_i, x_{i+1}),$ $\sigma$ is a path, $x_i\in \sigma$, $n$ is the DoF of the robotic arm, and $l$ is the length of the path. 

We compared our method with PRM, RRT*, and BIT*, where RRT* and BIT* only test the performance of finding the initial solution. In addition, to better compare the performance of these methods, we modify all methods to fit the test set and select the appropriate parameters for each method. For example, we have enhanced the ability of RRT* to construct long branches by using decreasing-length collision detection, and we have experimentally demonstrated that this ability can significantly reduce the number of collision detections and increase the success rate of RRT*. Fig. \ref{Fig5} shows the performance of our method.

As shown in Fig. \ref{Fig5}(a), The average number of collision detections in the GNN model is about 0.49\% of PRM, 94.43\% of BIT*, and 1.74\% of RRT*, which shows a significant reduction in comparison with PRM in $\mathbb{R}^7$. The reduction of collision detection results in a significant increase in overall planning speed.
Fig. \ref{Fig5}(b) and Fig. \ref{Fig5}(c) show that, in each condition, our model spends about 1.63 seconds in total and 0.193 seconds on planning, while PRM spends 20.47 seconds in total and 12.02 seconds on planning. The average planning time of our method is about 62 times shorter than that of PRM, accounting for about 12\% of the total time, while the planning time of PRM accounts for 59\% of the total planning time.

We also show the cost of the path of each method in Fig. \ref{Fig5}(d) Just as our model aims to find a path that preferentially avoids obstacles, the proposed method is not optimal in the cost of the path. The average cost of each test of PRM, BIT*, and RRT* is 13.473, 12.271, and 16.010, respectively, while the cost of our method is 19.452. 

Another concern is the robustness of the method, we use the success rate and variance to measure the robustness of the method. In Fig. \ref{Fig5}(e), the success rate of the proposed method to find the path in different numbers of sampling points is about 98\%, while PRM is 97\%, BIT* is 97\%, and RRT* is 98\%. Besides, our algorithm has the smallest variance in all performances, and the success rate can reach up to 99\% when the number of sampling points is 1000. In other words, our method has better performance and robustness in finding a feasible path, with a poor performance in finding the optimal path. The above features make the algorithm more suitable for scenarios that need to quickly solve a feasible path, rather than an optimal path.

\subsection{Real-world Implementations}
Since we use the 6D-Kinova robotic arm for real-world testing, we add an extra fixed degree of freedom to the end of the robotic arm to keep the model unchanged. We test the performance of our method and classical methods in $\mathbb{R}^6$ in the simulated random worlds before conducting physical experiments. The comparison of performance is shown in Fig. \ref{Fig7}.

As the model trained in $\mathbb{R}^7$, our method still improves collision detection times, planning time, and success rate compared to other algorithms in $\mathbb{R}^6$. Compared to PRM, the suggested model decreases collision detection by 99.7\%. Additionally, the planning time is reduced by about 98.6\% in contrast to BIT*. The path cost and success rate of the classical algorithm are comparable. The performance of our method verifies the feasibility of the physical experiment and the robustness of the model.

\begin{figure}[t]
    \centering
    \subfloat[]{\includegraphics[width=1.6in,height=1.5in]{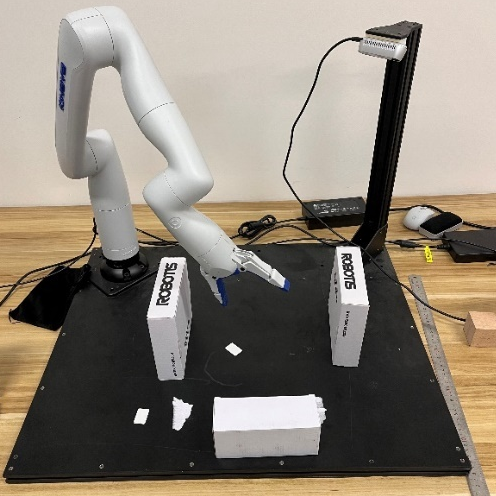}}
    \hfill
    \subfloat[]{\includegraphics[width=1.6in,height=1.5in]{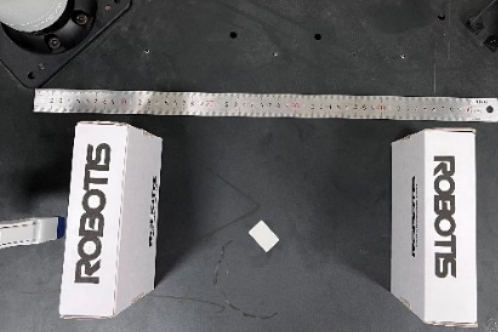}} 
    \hfill
    \subfloat[]{\includegraphics[width=1.6in,height=1.5in]{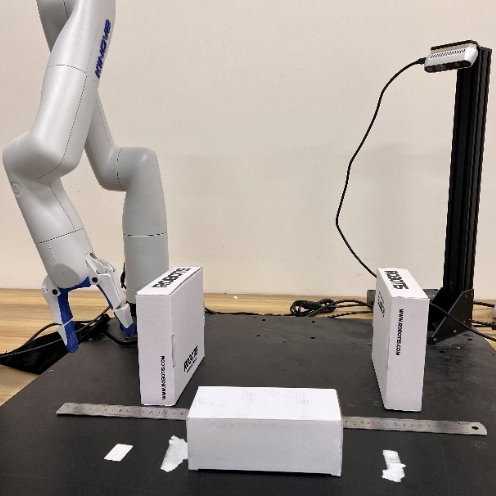}} 
    \hfill
    \subfloat[]{\includegraphics[width=1.6in,height=1.5in]{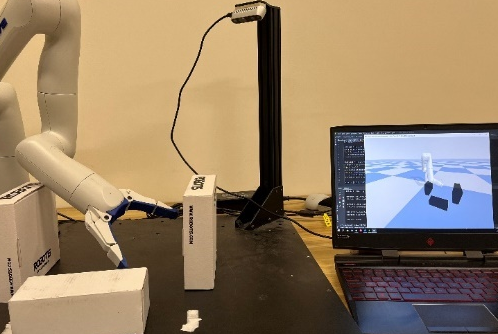}} 
    \caption{Experimental steps in a real-world test. From top left to bottom right is: (a) Experiment platform. (b) Measure the absolute position of obstacles. (c) Place obstacles randomly. (d) Get simulation results and do experiments.}
    \label{Fig6}
\end{figure}

As the proposed method cannot identify obstacles, it is necessary to manually measure the position and shape of obstacles. In each test, we first randomly place some obstacles (boxes) and measure the relative centroid position between the obstacles and the base of the robotic arm. Then we input the obstacles' information and use the proposed method to find a path and do a simulation. Finally, we convert the path into instructions and input it into the robotic arm to conduct real-world experiments. We record the successful planning of the path by method as successful planning and record whether the physical robotic arm is in continuous and non-repetitive motion as a suitable motion. The experimental steps are shown in Fig. \ref{Fig6}. 

At the beginning of testing, our method usually could not find a feasible solution. By analyzing the failure examples, we find that some of the mapped obstacles interfere with the initial or goal state, and make the model unable to calculate the guidance value set for such states (the same as the other interfered states), resulting in poor performance in such cases. Such a problem can be solved by increasing the resolution of the mapping space or using the original obstacle arrays as the input of the model. Since both of the above methods need to be solved by modifying the model, we avoid the problems by increasing the distance between the obstacle and the initial state and the goal state physically, which means reducing the resolution of the real space to fit the mapping space.

We test 10 experiments and successfully find the solution in 8 experiments. However, since these are not optimized, 2 of the paths are unsuitable for real robotic arms, which means that the paths are discontinuous or too close to obstacles. We also test BIT* in the same condition as the model, and BIT* successfully plans 9 times with 1 of the paths being unsuitable. In the real-world implementations, as we do not train the model with dynamics, our method performed poorly, with only about 60\% of the paths suitable for real robotic arms. Most unsuitable paths for robotic arms are prone to fall into local minimums, meaning that these paths will swing back and forth within an area,  which can be solved by increasing the proportion of path encoding.

\section{CONCLUSIONS AND FUTURE WORK}
\label{tit6}
In this paper, we present a learning-based planning method to guide the planner to choose the appropriate path. In particular, we design a GNN-based model to generate a weight called guidance value for each neighbor of a state and use the guidance value to search for paths. We evaluate the model in the simulated random worlds and the real-world implementations. In the simulation environment, the model shows a higher success rate, faster planning speed, and less collision detection in the high-dimensional environment. However, in the real-world environment, our method often fails to account for dynamics and therefore performs poorly on real robotic arms.

Compared to existing path planning methods, our model improves collision detection and planning time, but it comes at a loss in path cost and precision. In the future, we plan to add dynamics to the training of GNN models and reduce path costs by adding a new algorithm. Besides, we intend to further strengthen the proportion of path coding to alleviate the problem of the proposed method falling into local minimums. Additionally, we plan to modify the GNN model to gather obstacle information directly, as the computational and resolution of the mapping space decreases efficiency.

\addtolength{\textheight}{-2cm}

\end{document}